\documentclass[11pt]{article}

\usepackage[preprint]{acl}

\usepackage{times}
\usepackage{latexsym}

\usepackage[T1]{fontenc}

\usepackage[utf8]{inputenc}

\usepackage{microtype}

\usepackage{inconsolata}

\usepackage{graphicx}

\usepackage{array}
\usepackage{booktabs}
\usepackage{algorithm}
\usepackage{algorithmic}
\usepackage{natbib} 
\usepackage{caption} 
\usepackage{multirow}
\usepackage{amsmath}
\usepackage{makecell}
\usepackage{booktabs}
\usepackage[table]{xcolor}  
\usepackage{amssymb}  
\usepackage{longtable}
\usepackage{epstopdf}
\usepackage{graphicx}
\usepackage{times}  
\usepackage{helvet}  
\usepackage{courier}  
\usepackage{enumitem}

\title{Graph Topology Meets Energy-Driven Pathfinding: Preserving Contextual Consistency in Retrieval-Augmented Generation}

\author{
\textbf{Yufeng Li\textsuperscript{1}},
\textbf{Zaijing Li\textsuperscript{2}},
\textbf{Fuchen Zhang\textsuperscript{1}},
\textbf{Zongzong Wu\textsuperscript{1}},
\textbf{Ming Zhao\textsuperscript{1}},
\textbf{Fengxiao Tang\textsuperscript{1,*}}
\\
\textsuperscript{1}Central South University, Changsha, China \\
\textsuperscript{2}Harbin Institute of Technology, Shenzhen, China \\
\texttt{liyufeng@csu.edu.cn} \\
}

\begin{document}
\maketitle

\renewcommand{\thefootnote}{\fnsymbol{footnote}}
\footnotetext[1]{Corresponding author.}
\renewcommand{\thefootnote}{\arabic{footnote}}

\begin{abstract}
Retrieval-augmented generation (RAG) frequently integrates knowledge graphs (KGs) for complex multi-hop question answering. However, their performance is increasingly bottlenecked by the knowledge structuring and reasoning processes. (i) Severe semantic loss: Distilling continuous texts into discrete symbols strips away contextual nuances and narrative coherence. (ii) Pronounced modality mismatch: Fragmented graph structures clash with the continuous texts LLMs are pre-trained on, while rewriting them into natural language introduces generative noise and omits information. To address these limitations, we propose \textbf{ThreadRAG}, a semantic-topology-based RAG framework for context-preserving knowledge structuring and modality-native evidence reasoning. ThreadRAG introduces a \textbf{S}emantic \textbf{T}opology \textbf{M}apper (\textbf{STM}), which organizes retrieved documents into context-intact Semantic Nodes and constructs a \textbf{S}emantic \textbf{T}opology \textbf{G}raph (\textbf{STG}) through narrative, semantic, and lexical edges. It further incorporates \textbf{E}nergy-\textbf{D}riven \textbf{P}athfinding (\textbf{EDP}), which identifies query-relevant subgraphs via energy diffusion and deterministically linearizes them into coherent textual evidence paths. Experiments on three multi-hop QA benchmarks demonstrate that ThreadRAG consistently outperforms strong baselines. Notably, it achieves a 14.4\% EM improvement over text-centric methods on HotpotQA and surpasses the best baseline by 5.8\% EM on 2WikiMultihop, while reducing token cost to 23.3\% of the strongest baseline.
\end{abstract}

\section{Introduction}
Retrieval-augmented generation (RAG) has become a powerful paradigm for enhancing large language models (LLMs) on knowledge-intensive tasks by grounding generation in external evidence \cite{gao2023retrieval}. However, complex multi-hop question answering (QA) remains challenging for existing RAG systems \cite{lewis2020retrieval, izacard2022few, trivedi2023interleaving}. Unlike single-hop QA, multi-hop QA requires models to identify scattered evidence across multiple documents, connect them through latent relations, and perform multi-step reasoning \cite{shao2023enhancing}.

\begin{figure*}[t]
\centering
\vspace{-0.5cm}
\includegraphics[width=2.1\columnwidth]{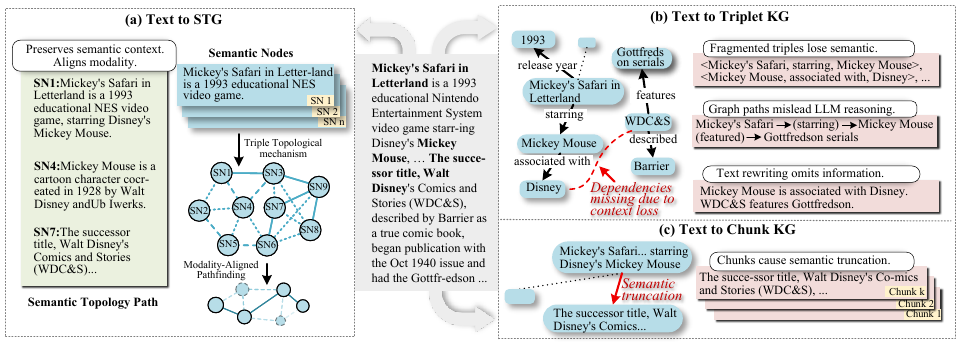}
\vspace{-0.8cm}
\caption{Comparison of graph-RAG paradigms. \textbf{Left}: Our proposed ThreadRAG (a) structures raw texts into a Semantic Topology Graph (STG) and extracts Semantic Topology Paths (ST-Paths), preserving semantic context and aligning modality for LLM deduction. \textbf{Right}: Existing paradigms, Triplet KG (b) and Chunk KG (c), fragment texts into discrete components, introducing severe semantic loss and modality mismatch.}
\vspace{-0.5cm}
\label{fig:intro}
\end{figure*}

Graph-enhanced RAG methods address this challenge by introducing structural priors over retrieved knowledge. Despite their promise, existing graph-RAG paradigms often suffer from a mismatch between how knowledge is structured and how LLMs perform reasoning. As illustrated in Figure~\ref{fig:intro}, this mismatch manifests in two key limitations. First, existing graph construction pipelines can incur substantial \emph{semantic loss}. Triplet-based knowledge graphs compress continuous text into rigid symbolic tuples, often discarding contextual nuances, while chunk-based graphs split documents into isolated segments, leading to semantic truncation and noisy boundaries. Such fragmentation weakens the continuous contextual pathways required for multi-hop reasoning. Second, existing graph representations introduce a \emph{modality mismatch} at inference time. LLMs are primarily optimized to consume coherent natural language sequences, whereas graph-RAG methods typically provide fragmented graph paths or rely on generative verbalization to convert subgraphs into text. These conversions can introduce redundancy, hallucinated transitions, or omission of critical evidence, thereby limiting the effectiveness of structural priors.

To address these limitations, we propose \textbf{ThreadRAG}, a semantic-topology-based RAG framework that preserves textual context during knowledge structuring and converts graph evidence into LLM-native reasoning sequences. ThreadRAG consists of two key components: a \textbf{S}emantic \textbf{T}opology \textbf{M}apper (\textbf{STM}) for high-fidelity knowledge structuring, and an \textbf{E}nergy-\textbf{D}riven \textbf{P}athfinding module (\textbf{EDP}) for query-aware evidence selection and topological linearization.

\textbf{A Semantic Topology Mapper for high-fidelity knowledge structuring.} 
To mitigate semantic loss introduced by conventional graph construction, STM structures retrieved documents into an LLM-oriented \textbf{S}emantic \textbf{T}opology \textbf{G}raph (\textbf{STG}). Instead of extracting symbolic triplets or relying on isolated chunks, STM organizes retrieved documents into self-contained \emph{Semantic Nodes} that preserve local textual context. It then connects these nodes through complementary topological relations, including narrative, semantic, and lexical edges. Narrative edges preserve the original textual order, while semantic and lexical edges capture cross-node associations beyond surface adjacency. This design allows STG to retain the semantic continuity of raw text while exposing the relational structure needed for evidence navigation. These relations transform fragmented retrieved evidence into a navigable topology without discarding the natural context required for multi-hop reasoning.

\textbf{An Energy-Driven Pathfinding module for modality-native knowledge representation.} 
To bridge the modality mismatch between discrete graph structures and sequence-based LLM reasoning, EDP converts query-relevant topology into continuous textual evidence paths. Given a query, EDP first identifies relevant Semantic Nodes as reasoning anchors, and then performs energy-driven propagation and dynamic pruning over the STG to obtain a compact, query-specific evidence subgraph. Instead of directly prompting the LLM with discrete graph paths or using free-form generation to verbalize the subgraph, EDP deterministically linearizes the selected nodes according to their topological and narrative structure. This process preserves the structural priors of the graph while presenting evidence in a form naturally compatible with LLM pretraining. The resulting \emph{Semantic Topology Paths} provide continuous, context-preserving, and sequence-native evidence, enabling more coherent and reliable multi-hop reasoning.

We conduct comprehensive evaluations across three multi-hop QA benchmarks. ThreadRAG consistently surpasses the best mainstream graph-enhanced paradigms across all three datasets (e.g., achieving a 3.4\% EM improvement over KnowTrace on HotpotQA). In graph representation evaluations, replacing traditional global graphs with our STG yields a 12.6\% EM improvement. Furthermore, our sequence-native STG-Paths strictly dominate all prevailing knowledge representation modalities, peaking at 0.588 EM on 2WikiMultihop. These results indicate that high-fidelity knowledge structuring and topological linearization significantly boost LLM deduction accuracy. Our contributions are as follows:

\begin{itemize}[leftmargin=*, itemsep=0pt, parsep=0pt, topsep=0pt]
    \item We propose a \textbf{S}emantic \textbf{T}opology \textbf{M}apper that reduces semantic loss by organizing retrieved text into a context-preserving \textbf{S}emantic \textbf{T}opology \textbf{G}raph.
    
    \item We introduce an \textbf{E}nergy-\textbf{D}riven \textbf{P}athfinding module that selects query-relevant evidence and linearizes discrete graph structures into sequence-native prompts for LLM reasoning.
    
    \item We present \textbf{ThreadRAG}, a unified graph-enhanced RAG framework that consistently improves multi-hop QA performance over strong baselines across three benchmarks.
\end{itemize}

\begin{figure*}[t]
\centering
\vspace{-0.5cm}
\includegraphics[width=2.1\columnwidth]{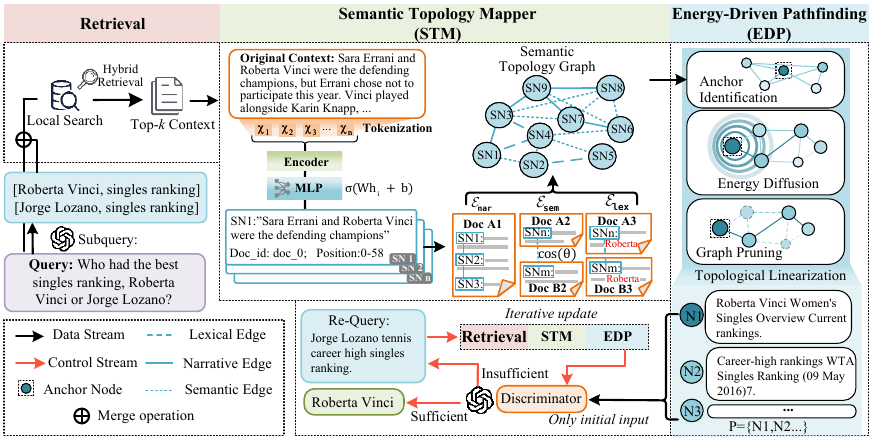}
\caption{Overview of ThreadRAG. Given a query and top-$k$ retrieved contexts, the STM structures raw text into STG. The EDP then isolates a query-relevant subgraph via energy diffusion and linearizes it into modality-native prompts. Finally, an adaptive agent uses these prompts to iteratively refine retrieval or synthesize answers.}
\vspace{-0.3cm}
\label{fig:ST-RAG}
\end{figure*}

\section{Related Work}

\textbf{Semantic Loss in Graph RAG.} Integrating KGs into RAG provides explicit structural priors for complex reasoning. Mainstream paradigms like GraphRAG \cite{edge2024local} extract rigid entity-relation triplets, which inadvertently strip away contextual nuances and cause severe semantic loss. To preserve semantics, early NLP QA approaches \cite{fang2020hierarchical, qiu2019dynamically, ding2019cognitive} retain original text spans. However, their heterogeneous node designs target GNN aggregation rather than LLM prompt reconstruction, irreversibly fragmenting the logical sequence. To better align with LLMs, recent methods such as HippoRAG \cite{gutierrez2024hipporag} utilize broader text-chunk nodes. While mitigating symbolic extraction loss, their unconstrained chunk assembly introduces semantic truncation and structural noise. Consequently, these paradigms suffer from semantic loss, depriving LLMs of the continuous contextual pathways essential for multi-hop reasoning. We address this bottleneck by proposing a topological framework that seamlessly embeds structural priors without compromising narrative integrity.

\textbf{Modality Alignment in Graph RAG.} A critical bottleneck in graph-augmented reasoning is the inherent modality gap between discrete graphs and continuous natural language. Several structural reasoning methods \cite{luo2024reasoning, sun2024think, tan2025paths} compel LLMs to process fragmented symbolic data over discrete entity-relation paths. This modality mismatch permanently strips away crucial contextual nuances like temporal constraints. To bridge this, generative rewriting approaches \cite{wu2024cotkr, sarthi2024raptor} employ LLMs to rewrite subgraphs into natural language, introducing substantial token overhead and hallucination risks. Conversely, chunk-as-node approaches \cite{pham2026discourse, yue2026litesemrag} preserve text spans but merely concatenate nodes as isolated chunks, failing to reconstruct cross-document narrative coherence. ThreadRAG addresses these limitations via a Topological Pathfinding mechanism. Using topological linearization, it restructures fragments into a coherent sequence. This aligns graph topologies with the LLM's native modality, enabling multi-hop comprehension without generative rewriting overhead.

\vspace{-0.1cm}
\section{ThreadRAG}
\vspace{-0.1cm}

In this section, we present the overall architecture of ThreadRAG (Figure \ref{fig:ST-RAG}). The framework primarily consists of an STM module that structures retrieved contexts into an STG (Sec. \ref{sec:Semantic Topology Mapper}), a EDP module that extracts and linearizes query-relevant subgraphs into continuous text sequences (Sec. \ref{sec:Energy-Driven Pathfinding}), and an adaptive reasoning agent that iteratively refines the evidence base to synthesize the final answer (Sec. \ref{sec:Adaptive Reasoning Agent}).

\subsection{Semantic Topology Mapper}
\label{sec:Semantic Topology Mapper}
Existing graph-RAG methods fragment continuous texts into discrete components, destroying narrative continuity and inducing severe semantic loss. Motivated by this, our \textbf{S}emantic \textbf{T}opology \textbf{M}apper (\textbf{STM}) extracts context-intact semantic nodes and interconnects them via narrative, semantic, and lexical edges into an STG. This fundamentally anchors structural reasoning priors directly into native text sequences, thereby transforming fragmented documents into a high-fidelity network.

\noindent\textbf{Semantic Node extraction}
Given a query $q$, an LLM extracts key entities $\mathcal{E}_k$ to construct a retrieval query set $\mathcal{Q}_{\text{ret}} = \{q\} \cup \mathcal{E}_k$. These queries are independently issued to a hybrid retrieval index over a local Wikipedia corpus, combining BM25 lexical matching with dense semantic retrieval using equal weights. The retrieved passages are pooled, deduplicated, and ranked according to the fused retrieval score to yield the top-$k$ context set:
\begin{equation}
\mathcal{D} = \operatorname{top\text{-}k} \bigg( \bigcup_{q_j \in \mathcal{Q}_{\text{ret}}} \operatorname{HybridRetrieve}(q_j) \bigg)
\end{equation}

For each document $d_i \in \mathcal{D}$, the raw text is tokenized into a sequence $X = \{x_1, x_2, \ldots, x_n\}$. This sequence is processed through a lightweight encoder to obtain contextualized hidden representations $\mathbf{h}_j \in \mathbb{R}^d$ for each token $x_j$. To identify semantic breakpoints, each representation $\mathbf{h}_j$ is mapped to a boundary probability $p_j \in [0,1]$ via a MLP(see Appendix \ref{sec:segmentation_details} for detailed configurations):
\begin{equation}
p_j = \sigma(\mathbf{W}_2(\operatorname{ReLU}(\mathbf{W}_1 \mathbf{h}_j + \mathbf{b}_1)) + b_2)
\end{equation}
where $\sigma$ denotes the sigmoid activation function, and $\mathbf{W}_1, \mathbf{W}_2, \mathbf{b}_1, b_2$ are learnable parameters. Tokens satisfying $p_j > \theta$, given a predefined threshold $\theta$, act as cut-off points. The continuous document is deterministically partitioned at these boundaries to yield a set of Semantic Nodes $\mathcal{V} = \{v_1, v_2, \ldots, v_m\}$. Each node $v_k \in \mathcal{V}$ is formally represented as a tuple:
\begin{equation}
v_k = (t_k, d_k, s_k, \tau_k)
\end{equation}
where $t_k$ is the raw natural language text span, $d_k$ is the document identifier, and $[s_k, \tau_k)$ denotes the exact character-level position interval of $t_k$ within $d_k$. Compared to existing LLM-based extraction paradigms, this feed-forward mechanism achieves context-intact segmentation with millisecond-level latency and strictly avoids generative noise.

\noindent\textbf{Semantic Topology Graph construction}
Given the node set $\mathcal{V}$, the STG is completed by constructing three complementary types of edges that capture distinct relational signals between SNs. Each edge type encodes a different dimension of inter-node relatedness, and their weights are fused under a unified formulation.

\textit{Narrative edges} $\mathcal{E}_{\text{nar}}$ connect positionally adjacent nodes within the same document to preserve the original reading order. Nodes sharing $d_i$ are sorted by start position $s_j$, with a directed edge $(v_j, v_{j+1}) \in \mathcal{E}_{\text{nar}}$ added between consecutive pairs.

\textit{Semantic edges} $\mathcal{E}_{\text{sem}}$ capture cross-document topical relatedness. Each node $v_i$ is represented by a dense embedding $\mathbf{h}_i$, connecting top-$k$ cross-document pairs $(d_i \neq d_j)$ with weight:
\begin{equation}
\text{Sim}_{\text{sem}}(v_i, v_j) = \frac{\mathbf{h}_i \cdot \mathbf{h}_j}{\|\mathbf{h}_i\|\,\|\mathbf{h}_j\|}
\end{equation}

\textit{Lexical edges} $\mathcal{E}_{\text{lex}}$ capture explicit entity co-reference. For named entities $\mathcal{N}_i$ extracted from $v_i$, cross-document pairs sharing entities are connected via Jaccard similarity:
\begin{equation}
\text{Sim}_{\text{lex}}(v_i, v_j) = \frac{|\mathcal{N}_i \cap \mathcal{N}_j|}{|\mathcal{N}_i \cup \mathcal{N}_j|}
\end{equation}

The final edge weight $W(v_i, v_j)$ balances local narrative continuity against global cross-document association:
\begin{equation}
\begin{split}
&W(v_i, v_j) = (1-\gamma)\bigl(\alpha \cdot \text{Sim}_{\text{sem}}(v_i, v_j) \\
&\quad + \beta \cdot \text{Sim}_{\text{lex}}(v_i, v_j)\bigr) + \gamma \cdot \mathbf{1}_{\text{nar}}(v_i, v_j)
\end{split}
\end{equation}
where $\mathbf{1}_{\text{nar}}(v_i, v_j) = 1$ iff $(v_i, v_j) \in \mathcal{E}_{\text{nar}}$, $\gamma \in (0,1)$ scales the narrative prior, and $\alpha + \beta = 1$. The complete edge set $\mathcal{E} = \mathcal{E}_{\text{nar}} \cup \mathcal{E}_{\text{sem}} \cup \mathcal{E}_{\text{lex}}$ alongside $\mathcal{V}$ defines the STG $\mathcal{G} = (\mathcal{V}, \mathcal{E})$.

\subsection{Energy-Driven Pathfinding}
\label{sec:Energy-Driven Pathfinding}
Current methods relying on fragmented graph components or generative rewriting struggle with a modality mismatch and suffer from generative noise or critical information omission. Our \textbf{E}nergy-\textbf{D}riven \textbf{P}athfinding (\textbf{EDP}) module addresses this by executing a context-preserving pathfinding mechanism over the STG. By isolating a query-specific subgraph and deterministically reconstructing it into continuous textual sequences, EDP overcomes structural fragmentation via anchor-driven energy diffusion and topological linearization.

\noindent\textbf{Anchor Identification.} Given the STG $\mathcal{G} = (\mathcal{V}, \mathcal{E})$ and a query $q$, the pathfinding stage begins by identifying a set of anchor nodes as entry points. Each node $v_i$ is scored by combining its semantic relevance to $q$ with an entity-level matching signal:
\vspace{-0.2cm}
\begin{equation}
s_i = \text{Sim}_{\text{sem}}(\mathbf{h}_i, \mathbf{h}_q) + \lambda \cdot \max_{e_j \in \mathcal{E}_k} \text{Sim}_{\text{sem}}(\mathbf{h}_i, \mathbf{h}_{e_j})
\end{equation}
where $\mathbf{h}_q$ and $\mathbf{h}_{e_j}$ are the embeddings of the query and extracted entities, respectively, and $\lambda$ is the entity boost weight. The top-$K$ nodes ranked by $s_i$ form the anchor set $\mathcal{A} \subseteq \mathcal{V}$.

\noindent\textbf{Energy Diffusion.} With the anchor set established, we construct an initial energy vector $\mathbf{r}$. Each anchor is assigned an energy proportional to its score, modulated by a provenance weight $w_i$ that prioritizes locally retrieved nodes:
\vspace{-0.1cm}
\begin{equation} 
r_i = \begin{cases} s_i \cdot w_i \cdot (1 + \mu \cdot b_i) & \text{if } v_i \in \mathcal{A} \\ 0 & \text{otherwise} \end{cases} 
\end{equation}
\begin{equation} 
b_i = \max_{e_j \in \mathcal{E}_k}\, \text{Sim}_{\text{sem}}(\mathbf{h}_i, \mathbf{h}_{e_j})
\end{equation}
where $\mu$ controls the entity bonus strength. After $\ell_1$-normalization of $\mathbf{r}$, energy is propagated across the STG via Personalized PageRank (PPR). The stationary distribution $\mathbf{p}$ satisfies:
\begin{equation}
\mathbf{p} = (1-\alpha)\mathbf{W}\mathbf{p} + \alpha\mathbf{r}
\end{equation}
where $\mathbf{W}$ is the row-normalized adjacency matrix weighted by the fused edge weights, and $\alpha \in (0,1)$ is the restart probability. This diffusion process allows scattered evidence to organically accumulate in a high-energy topological subregion.

\noindent\textbf{Graph Pruning.} Upon convergence, nodes with stationary energy $p_i$ falling below a threshold $\epsilon$ are pruned. A connectivity constraint is then enforced via a BFS traversal rooted at $\mathcal{A}$, retaining only reachable nodes to produce a compact, topologically grounded subgraph $\mathcal{G}' = (\mathcal{V}', \mathcal{E}')$. The retained nodes $\mathcal{V}'' \subseteq \mathcal{V}'$ (after deduplication and budget filtering) are subsequently deterministically linearized. We sort these nodes based on their physical coordinates $(d_i, s_i)$, ordering them primarily by document retrieval relevance and secondarily by intra-document narrative sequence. The final modality-native context is assembled as:
\begin{equation}
C = \operatorname{concat}(t_i \mid v_i \in \mathcal{V}'')
\end{equation}
This linearization operation effectively unravels the discrete pruned graph back into a continuous textual sequence, providing the downstream LLM with a contextually intact and strictly modality-aligned prompt.

\subsection{Adaptive Reasoning Agent}
\label{sec:Adaptive Reasoning Agent}
The linearized context $C$ obtained from topological pathfinding is passed to a reasoning agent that iteratively refines its evidence base until the query can be answered with sufficient confidence. 

At each iteration, a sufficiency discriminator evaluates the current context $C$ against the query $q$. If $C$ is deemed sufficient, the agent proceeds directly to answer synthesis; otherwise, the discriminator outputs the missing information as structured knowledge triples, which are forwarded to the sub-query generator to produce targeted retrieval entities. These entities are issued as new queries to the hybrid index, and the retrieved documents are processed through Steps 2–4 of the ThreadRAG pipeline—segmentation, graph augmentation, and topological pathfinding—to yield a refined context $C'$ that replaces $C$ for the next iteration.

This loop repeats for at most $N$ iterations. Upon termination, the final context is passed to an answer synthesis module that reasons step by step and produces a concise answer enclosed in a structured output tag, from which the final answer string is extracted. The overall procedure is summarized in Algorithm \ref{alg:inference}.

\vspace{-0.2cm}
\begin{algorithm}[H]
\caption{Inference Process of ThreadRAG}
\label{alg:inference}
\textbf{Require:} LLM $M$; BM25 index over corpus; max iterations $N$; energy threshold $\epsilon$; MLP threshold $\theta$ \\
\textbf{Input:} question $q$ \\
\textbf{Output:} linearized context $C$; final answer $a$

\begin{algorithmic}[1]
\STATE \textcolor{blue}{// Phase 1: Semantic Topology Mapper}
\STATE $\mathcal{E}_k \leftarrow M_{\text{extract}}(q)$; \quad $\mathcal{Q}_{\text{ret}} \leftarrow \{q\} \cup \mathcal{E}_k$ 
\STATE $\mathcal{D} \leftarrow \text{top-}k \left( \bigcup_{q_j \in \mathcal{Q}_{\text{ret}}} \text{HybridRetrieve}(q_j) \right)$ 
\STATE \textcolor{blue}{// Yields $v_i = (t_i, d_i, s_i, \tau_i)$}
\STATE $\mathcal{V} \leftarrow \text{MLP\_Segmentation}(\mathcal{D}, \theta)$ \quad 
\STATE $\mathcal{E} \leftarrow \mathcal{E}_{\text{nar}} \cup \mathcal{E}_{\text{sem}} \cup \mathcal{E}_{\text{lex}}$; \quad $\mathcal{G} \leftarrow (\mathcal{V}, \mathcal{E})$ 

\STATE \textcolor{blue}{// Phase 2: Energy-Driven Pathfinding}
\STATE $\mathcal{A} \leftarrow \text{Identify\_Anchors}(\mathcal{G}, q, \mathcal{E}_k)$ 
\STATE $\mathbf{p} \leftarrow \text{PPR\_Diffusion}(\mathcal{G}, \text{Init\_Energy}(\mathcal{A}))$
\STATE $\mathcal{V}' \leftarrow \left\{ v_i \in \mathcal{V} \mid p_i \geq \epsilon \right\}$ 
\STATE $\mathcal{G}' \leftarrow \text{BFS\_Traversal}(\mathcal{V}', \mathcal{A})$ 
\STATE $C \leftarrow \operatorname{concat}(\text{Sort\_by\_Coordinates}(v_i \in \mathcal{G}'))$ 

\STATE \textcolor{blue}{// Phase 3: Adaptive Reasoning Agent}
\FOR{$iter=1$ \textbf{to} $N$} 
    \STATE $\{\text{FLAG}, \mathcal{E}_{\text{miss}}\} \leftarrow M_{\text{eval}}(C, q)$ 
    \IF{FLAG} \STATE \textbf{break} \ENDIF
    \STATE $\mathcal{D}_{\text{new}} \leftarrow \text{HybridRetrieve}(\mathcal{E}_{\text{miss}})$
    \STATE $\mathcal{G} \leftarrow \text{STM\_Augment}(\mathcal{G}, \mathcal{D}_{\text{new}})$
    \STATE $C \leftarrow \text{EDP\_Linearize}(\mathcal{G}, q)$
\ENDFOR

\STATE $a \leftarrow M_{\text{synth}}(C, q)$ \\
\STATE \textcolor{blue}{// Final Answer Synthesis}
\STATE \textbf{return} $C, a$
\end{algorithmic}
\end{algorithm}

\vspace{-0.5cm}
\begin{table*}[ht!]
\centering
\resizebox{\textwidth}{!}{
\begin{tabular}{l | cc cc cc | cc cc cc | cc cc cc}
\toprule

\multirow{3}{*}{\textbf{Methods}} & 
\multicolumn{6}{c|}{\cellcolor{gray!15} LLaMA3-8B-Instruct} & 
\multicolumn{6}{c|}{\cellcolor{gray!15} Qwen3.5 plus} & 
\multicolumn{6}{c}{\cellcolor{gray!15} GPT-4o mini} \\

\cmidrule{2-19}
& \multicolumn{2}{c}{\textbf{HotpotQA}} & \multicolumn{2}{c}{\textbf{2Wiki}} & \multicolumn{2}{c|}{\textbf{MuSiQue}}
& \multicolumn{2}{c}{\textbf{HotpotQA}} & \multicolumn{2}{c}{\textbf{2Wiki}} & \multicolumn{2}{c|}{\textbf{MuSiQue}}
& \multicolumn{2}{c}{\textbf{HotpotQA}} & \multicolumn{2}{c}{\textbf{2Wiki}} & \multicolumn{2}{c}{\textbf{MuSiQue}} \\

\cmidrule(lr){2-3} \cmidrule(lr){4-5} \cmidrule(lr){6-7} 
\cmidrule(lr){8-9} \cmidrule(lr){10-11} \cmidrule(lr){12-13}
\cmidrule(lr){14-15} \cmidrule(lr){16-17} \cmidrule(lr){18-19}
& EM & F1 & EM & F1 & EM & F1 & EM & F1 & EM & F1 & EM & F1 & EM & F1 & EM & F1 & EM & F1 \\
\midrule
\multicolumn{19}{c}{\textit{Non-Graph RAG}} \\
\midrule
RA-CoT 
& .204 & .330 & .202 & .261 & .142 & .210 
& .312 & .429 & .268 & .372 & .162 & .249 
& .310 & .425 & .264 & .359 & .158 & .243 \\

ReAct  
& .268 & .379 & .244 & .335 & .216 & .311 
& .358 & .476 & .342 & .487 & .248 & .378 
& .349 & .475 & .338 & .483 & .242 & .383 \\

IRCoT  
& .318 & .430 & .288 & .367 & .224 & .327 
& .442 & .565 & .374 & .519 & .262 & .376
& .438 & .559 & .372 & .496 & .258 & .373 \\

Self-Ask 
& .256 & .373 & .222 & .331 & .190 & .281 
& .356 & .471 & .326 & .466 & .204 & .325 
& .348 & .465 & .328 & .471 & .200 & .317 \\

Iter-RetGen 
& .288 & .388 & .268 & .351 & .231 & .315 
& .424 & .534 & .370 & .503 & .250 & .373 
& .418 & .529 & .366 & .489 & .252 & .374 \\

\midrule
\multicolumn{19}{c}{\textit{Graph RAG}} \\
\midrule
HippoRAG 
& .324 & .403 & .318 & .411 & .202 & .274 
& .418 & .557 & .466 & .569 & .234 & .333 
& .412 & .549 & .458 & .572 & .219 & .327 \\

LightRAG 
& .286 & .357 & .288 & .365 & .182 & .249 
& .328 & .429 & .342 & .424 & .204 & .303 
& .312 & .402 & .326 & .429 & .194 & .293 \\

CoTKR 
& .304 & .367 & .312 & .366 & .212 & .309 
& .354 & .411 & .368 & .451 & .226 & .321 
& .368 & .435 & .376 & .487 & .234 & .303 \\

EFSUM 
& .264 & .343 & .266 & .351 & .206 & .271 
& .322 & .417 & .294 & .386 & .224 & .297 
& .320 & .411 & .316 & .409 & .218 & .279 \\

ToG 
& .318 & .401 & .284 & .355 & .202 & .317 
& .446 & .557 & .456 & .521 & .243 & .351 
& .448 & .565 & .434 & .512 & .258 & .367 \\

PoG 
& .336 & .426 & .312 & .380 & .230 & .321 
& .516 & .639 & .518 & .601 & .282 & .371 
& .508 & .623 & .492 & .587 & .278 & .366 \\

ERA-CoT 
& .354 & .439 & .304 & .371 & .253 & .350 
& .478 & .613 & .442 & .553 & .278 & .385 
& .464 & .608 & .436 & .538 & .270 & .379 \\

knowtrace
& \underline{.396} & \underline{.486} & \underline{.338} & \underline{.413} & \underline{.292} & \underline{.379}
& \underline{.552} & \underline{.695} & \underline{.546} & \underline{.638} & \underline{.318} & \underline{.414} 
& \underline{.534} & \underline{.679} & \underline{.528} & \underline{.629} & \underline{.320} & \underline{.413} \\

\midrule
\rowcolor{blue!10}
ThreadRAG
& \textbf{.422} & \textbf{.513} & \textbf{.372} & \textbf{.453} & \textbf{.324} & \textbf{.417} 
& \textbf{.586} & \textbf{.704} & \textbf{.596} & \textbf{.671} & \textbf{.342} & \textbf{.453} 
& \textbf{.576} & \textbf{.685} & \textbf{.586} & \textbf{.663} & \textbf{.328} & \textbf{.437} \\

\bottomrule
\end{tabular}
}
\caption{Main results of multi-hop QA datasets. Best results are in \textbf{bold} and second best are \underline{underlined}.}
\label{tab:main_results}
\end{table*}

\section{Experiments}

\subsection{Experimental Setup}

\textbf{Datasets \& Metrics.}We evaluate ThreadRAG on three standard open-domain multi-hop QA benchmarks: HotpotQA \cite{yang2018hotpotqa}, 2WikiMultihopQA \cite{ho2020constructing}, and MuSiQue \cite{trivedi2022musique}. We use the same data splits as previous works \cite{trivedi2023interleaving, li2023leveraging} for evaluation. Following IRCoT \cite{trivedi2023interleaving}, our retrieval corpus aggregates all supporting and distractor passages to simulate a realistic open-domain setting. For evaluation, we report Exact Match (EM) and F1 scores to measure absolute correctness and token-level overlap, applying standard text normalization to both predictions and ground-truth answers prior to computation.

\textbf{Baselines.} We evaluate ThreadRAG against five categories of baselines: text-centric multi-hop methods reasoning purely over plain text (RA-CoT \cite{wei2022chain}, ReAct \cite{yao2022react}, IRCoT \cite{trivedi2023interleaving}, Self-Ask \cite{press2023measuring}, Iter-RetGen \cite{shao2023enhancing}), KG-Chunk RAG methods treating text chunks as nodes (HippoRAG \cite{gutierrez2024hipporag}, LightRAG \cite{guo2024lightrag}), KG-Rewriting RAG methods converting subgraphs into natural language (CoTKR \cite{wu2024cotkr}, EFSUM \cite{ko2024evidence}), KG-Path RAG methods reasoning along discrete structural paths (ToG \cite{sun2024think}, PoG \cite{tan2025paths}), and KG-Triplet RAG methods prompting with discrete relational triplets (ERA-CoT \cite{liu2024era}, KnowTrace \cite{li2025knowtrace}). For fair comparison, we adapt ToG and PoG to use the same local KG as LightRAG, replacing their external KGs.

\textbf{Implementation Details.} We use Llama-3-8B \cite{grattafiori2024llama}, Qwen-3.5-Plus \cite{yang2025qwen3}, and GPT-4o-mini \cite{hurst2024gpt} with \textit{nomic-embed-text} for embeddings \cite{nussbaum2024nomic}. Chunk size is 1200 characters with 150 overlap. We retrieve top-5 global contexts via Elasticsearch using equal BM25 and dense weights. STG settings are: 5-NN, 0.5/0.3 semantic/lexical thresholds, and edge weights $\gamma=0.3$, $\alpha=0.35$, $\beta=0.65$. Pathfinding parameters are: 3 anchors, 0.5 PPR restart, 0.01 pruning threshold, and 3 max iterations. All experiments run on 4 NVIDIA A6000 GPUs.

\subsection{Inference Performance Comparison}
Table \ref{tab:main_results} summarizes the main experimental results across three multi-hop QA datasets. First, we observe that KG-Chunk methods (e.g., HippoRAG) outperform KG-Rewriting approaches to a certain extent (e.g., CoTKR, achieving 0.324 vs. 0.304 EM on HotpotQA with LLaMA3-8B-Instruct). We attribute this to the generative noise and hallucinations introduced when rewriting discrete subgraphs into text. Second, iterative KG-Triplet prompting (e.g., KnowTrace) significantly surpasses KG-Path methods (e.g., PoG, 0.396 vs. 0.336 EM). This indicates that faulty path retrieval can easily inject erroneous logical chains, which LLMs struggle to verify or reject due to the severe loss of surrounding narrative context in discrete nodes.

Beyond these flawed paradigms, ThreadRAG directly grounds structural guidance within native text sequences to achieve state-of-the-art results. For example, it outperforms the strongest  baseline KnowTrace, by up to 5.8\% EM on 2Wiki (GPT-4o mini). Furthermore, it surpasses the text-centric IRCoT by a substantial 14.4\% EM on the complex HotpotQA dataset (Qwen3.5 plus). These significant gains demonstrate the effectiveness of our STG in providing high-fidelity structural priors. By providing explicit reasoning pathways without fragmenting the original narrative flow, ThreadRAG bridges the modality gap, bypassing both the semantic loss of symbolic extraction and the hallucination risks of generative rewriting.

\subsection{Ablation Study}

\begin{table}[t]
\centering
\resizebox{\columnwidth}{!}{
\begin{tabular}{ccc | ccc}
\toprule
\multicolumn{3}{c|}{Ablation Setting} & \multicolumn{3}{c}{Datasets} \\
\cmidrule(lr){1-3} \cmidrule(lr){4-6}
SN & STG & EDP & HotpotQA & 2Wiki & MuSiQue \\
\midrule
\rowcolor{blue!10}
\checkmark & \checkmark & \checkmark & \textbf{.422} & \textbf{.372} & \textbf{.324} \\
& & & .238 {\scriptsize \textcolor{red}{($\downarrow$ 18.4\%)}} & .224 {\scriptsize \textcolor{red}{($\downarrow$ 14.8\%)}} & .172 {\scriptsize \textcolor{red}{($\downarrow$ 15.2\%)}} \\
& & \checkmark & .326 {\scriptsize \textcolor{red}{($\downarrow$ 9.6\%)}} & .308 {\scriptsize \textcolor{red}{($\downarrow$ 6.4\%)}} & .232 {\scriptsize \textcolor{red}{($\downarrow$ 9.2\%)}} \\
\checkmark & \checkmark & & .398 {\scriptsize \textcolor{red}{($\downarrow$ 2.4\%)}} & .334 {\scriptsize \textcolor{red}{($\downarrow$ 3.8\%)}} & .286 {\scriptsize \textcolor{red}{($\downarrow$ 3.8\%)}} \\
& \checkmark & \checkmark & .388 {\scriptsize \textcolor{red}{($\downarrow$ 3.4\%)}} & .342 {\scriptsize \textcolor{red}{($\downarrow$ 3\%)}} & .292 {\scriptsize \textcolor{red}{($\downarrow$ 3.2\%)}} \\
\bottomrule
\end{tabular}
}
\caption{Ablation study of the STM (SN, STG) and EDP modules across three datasets.}
\label{tab:ablation}
\end{table}

Many unexplored questions surround the optimal knowledge representation and reasoning modalities for multi-hop QA. In this section, we conduct an extensive ablation study to validate the core components of ThreadRAG and summarize key findings.

\textbf{The STM preserves crucial contextual integrity.} As shown in Table \ref{tab:ablation}, removing the entire STM module (i.e., SN and STG) and using LLM-extracted triplets leads to a severe decline in performance, with the EM score dropping by 9.2\% on MuSiQue. Furthermore, replacing our MLP-based Semantic Nodes (SN) with fixed-length chunks results in a noticeable 3.4\% drop on HotpotQA. We attribute this to the fact that traditional triplet extraction or arbitrary chunking inevitably destroys the continuous narrative flow, resulting in irreversible semantic loss that hinders the LLM's ability to deduce complex answers.

\textbf{EDP effectively bridges the modality gap.} As shown in Table \ref{tab:ablation}, feeding the discrete STG nodes directly into the LLM without the EDP module's topological linearization causes a noticeable decline, such as an 3.8\% decrease on the MuSiQue dataset. We attribute this to the inherent modality mismatch between discrete graph structures and the LLM's native continuous language space. Without EDP to reconstruct the narrative sequence, the LLM struggles to effectively interpret the fragmented subgraphs, which introduces reasoning noise and degrades overall performance.

\subsection{More Analysis}

\begin{figure*}[t]
\centering
\includegraphics[width=\textwidth]{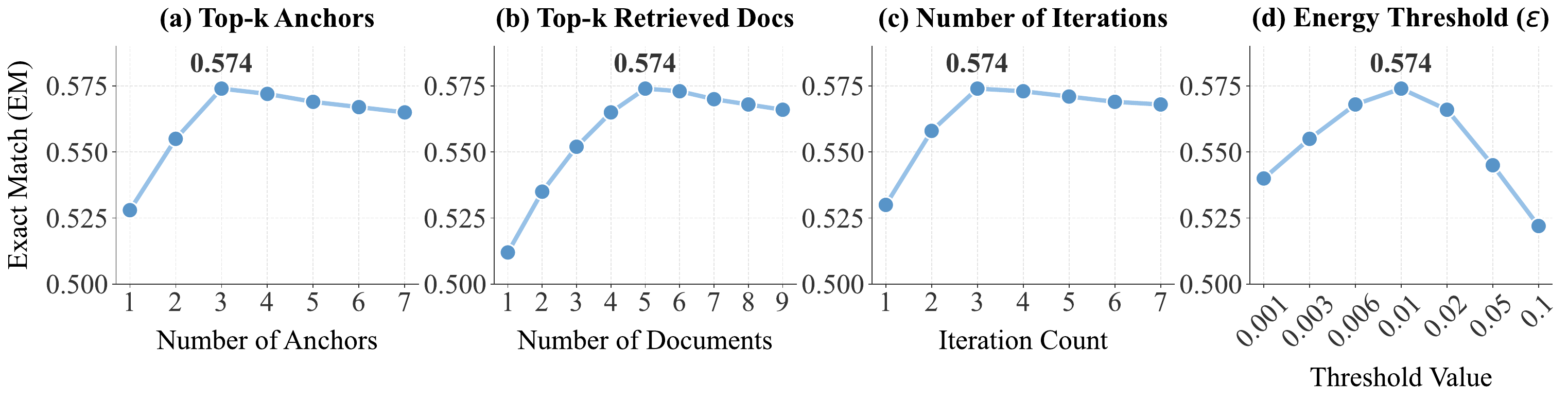}
\vspace{-0.9cm}
\caption{Hyperparameter sensitivity analysis on the HotpotQA dataset.}
\vspace{-0.3cm}
\label{fig:hyperparameters}
\end{figure*}

\textbf{Cost Analysis.}
To quantify the computational overhead, we measure token consumption across 100 randomly sampled queries per dataset using the \texttt{tiktoken} library (GPT-4o-mini standard). As shown in Table \ref{tab:token_costs}, ThreadRAG demonstrates a decisive advantage in computational efficiency by fundamentally decoupling structural knowledge generation from expensive LLM pre-computation across all datasets. In contrast, baseline methods like LightRAG and KnowTrace consume massive token budgets—up to 36.4M and 220K tokens, respectively—merely to build their graph structures. Furthermore, when evaluating the overall pipeline, ThreadRAG operates highly efficiently: its total token consumption accounts for merely 23.3\% to 26.3\% of the overhead incurred by the strongest baseline, KnowTrace. This confirms that ThreadRAG effectively resolves the prohibitive I/O bottlenecks inherent in traditional graph-enhanced RAG paradigms.

\begin{table}[t]
\centering
\resizebox{\columnwidth}{!}{
\begin{tabular}{l | ccc | ccc}
\toprule
\multirow{2}{*}{\textbf{Method}} & \multicolumn{3}{c|}{\textbf{HotpotQA}} & \multicolumn{3}{c}{\textbf{MuSiQue}} \\
\cmidrule(lr){2-4} \cmidrule(lr){5-7}
& Const. & Inf. & Total & Const. & Inf. & Total \\
\midrule
LightRAG & 26.2M & 47.4k & 26.25M & 36.4M & 57.6k & 36.46M \\
KnowTrace & 0.15M & 72.6k & 222.6k & 0.22M & 94.4k & 314.4k \\
\midrule
\rowcolor{blue!10}
ThreadRAG & \textbf{0} & 58.5k & \textbf{58.5k} & \textbf{0} & 73.4k & \textbf{73.4k} \\
\bottomrule
\end{tabular}
}
\caption{Comparison of construction (Const.), inference (Inf.), and total token costs.}
\label{tab:token_costs}
\end{table}

\textbf{Hyperparameter analysis.} To evaluate framework robustness, we conduct a hyperparameter analysis on the HotpotQA dataset using the GPT-4o-mini model. As illustrated in Figure \ref{fig:hyperparameters}, we investigate four key variables: initial anchors, retrieved documents, reasoning iterations, and the energy pruning threshold ($\epsilon$). ThreadRAG demonstrates remarkable stability, consistently maintaining Exact Match (EM) scores above 0.512 across all extreme settings. Performance optimally peaks at an EM of 0.574 with 3 anchors, 5 documents, 3 iterations, and a 0.01 threshold. Notably, exceeding these values yields marginal diminishing returns rather than sharp performance drops. This confirms ThreadRAG effectively balances computational efficiency with reasoning accuracy, avoiding unnecessary over-exploration.

\begin{figure}[t]
\centering
\includegraphics[width=0.5\textwidth]{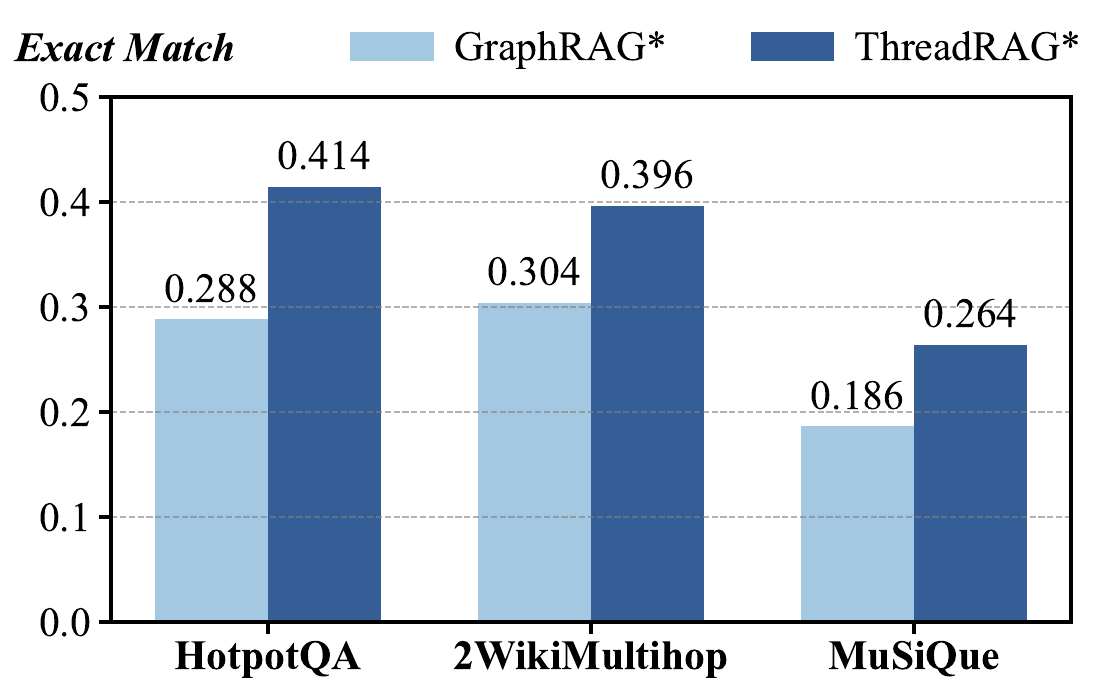}
\caption{Performance comparison of EM scores using different underlying graph representations.}
\vspace{-0.5cm}
\label{fig:graph_effect}
\end{figure}

\textbf{Graph Representation Comparison.} This experiment compares the conventional GraphRAG paradigm, which relies on the strict extraction of rigid entity-relation triplets, against our STG paradigm, which inherently preserves native semantic context and narrative flow. To facilitate a rigorous evaluation, we implement two specific variants: \textbf{GraphRAG$^\ast$}, which utilizes a pre-constructed global KG, and \textbf{ThreadRAG$^\ast$}, which adopts our proposed Semantic Topology Graph. All other components, including retrieval indices and prompting templates, are kept strictly identical. 

As illustrated in Figure \ref{fig:graph_effect}, ThreadRAG$^\ast$ consistently achieves superior performance across all benchmarks, securing substantial EM improvements ranging from 7.8\% to 12.6\%. Notably, on the HotpotQA dataset, our approach yields a significant 12.6\% improvement in EM (from 0.288 to 0.414). We attribute these substantial gains to STG's ability to preserve rich semantic narrative flows and ensure precise modality alignment via topological paths. In contrast, the discrete, symbolic triplets employed by GraphRAG$^\ast$ tend to fragment the original context, leading to inevitable semantic loss during complex multi-hop reasoning.

\begin{figure}[t]
\centering
\includegraphics[width=0.48\textwidth]{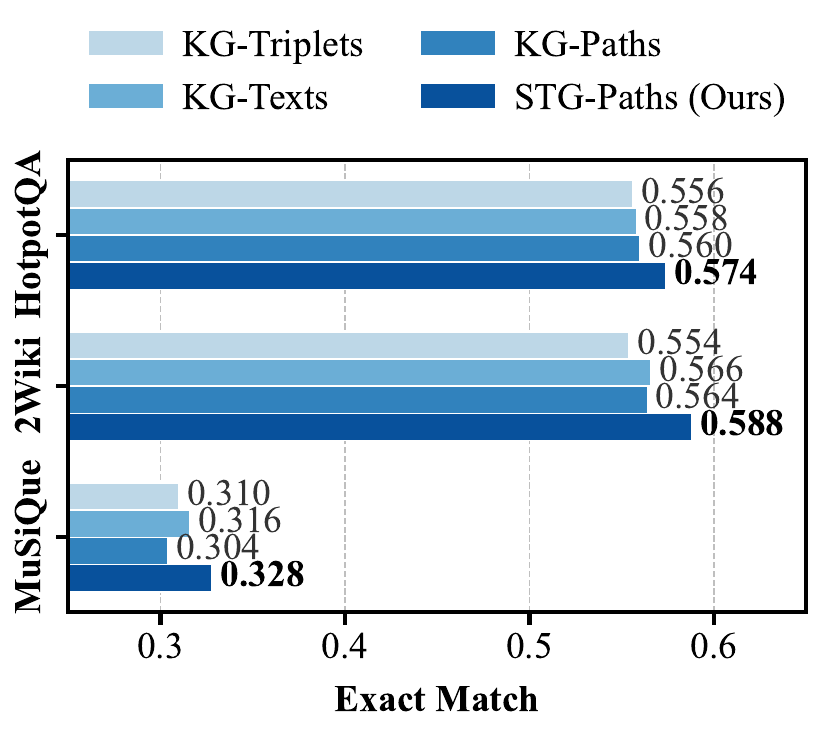}
\caption{Performance comparison of various knowledge representation strategies.}
\vspace{-0.5cm}
\label{fig:prompting_strategy}
\end{figure}

\textbf{knowledge representation strategies.} Since a critical concern lies in how to optimally format retrieved knowledge for LLM inference, we investigate four knowledge representation strategies: KG-Triplets, KG-Paths, KG-Texts, and our STG-Paths, respectively corresponding to discrete relational triplets, linear structural paths, LLM-rewritten natural language, and topological text paths. As illustrated in Figure \ref{fig:prompting_strategy}, STG-Paths consistently achieves the highest performance. We observe that forcing an LLM to rewrite triplets into natural language (KG-Texts) occasionally underperforms even naive KG-Triplets, which we attribute to hallucinations and semantic distortions introduced during rewriting. Conversely, while KG-Triplets and KG-Paths avoid rewriting risks, they suffer from severe semantic fragmentation and lack deep structural context. In stark contrast, STG-Paths resolves this dilemma. By directly extracting topological text paths, it preserves undistorted raw information while injecting structural priors. This dual advantage of strict semantic fidelity and logical coherence renders STG-Paths the most effective representation for our framework.

\section{Conclusion}
In this paper, we propose ThreadRAG to tackle complex multi-hop reasoning challenges. ThreadRAG integrates a Semantic Topology Graph (STG) for knowledge structuring and a Modality-Aligned Reasoning Mechanism via Energy-Driven pathfinding. As a core contribution, ThreadRAG linearizes discrete graph structures into continuous natural language sequences, effectively bridging the modality gap and preserving crucial narrative context. Extensive experimental results demonstrate that ThreadRAG achieves superior performance across multiple benchmarks, surpassing existing state-of-the-art baselines with low computational overhead.

\section*{Limitations}
Our work has limitations in certain scenarios. First, ThreadRAG relies on capturing structural priors through narrative, semantic, and lexical edges, so its performance improvement may not be significant in tasks with extremely sparse document links or where cross-document evidence chains are virtually non-existent, such as single-document standard QA tasks. Additionally, due to the dynamic nature of text segmentation and the complexity of multi-hop paths, even after energy diffusion and graph pruning, there may still be a few long-range contextual dependencies that the pathfinding mechanism fails to perfectly linearize in the exact chronological or logical order. This could result in minor narrative fragmentation during prompt reconstruction, affecting the reasoning accuracy of downstream LLMs. In future work, we aim to address these two issues. We plan to explore whether the model can effectively optimize the topological path selection and conduct a more fine-grained alignment of narrative flows to further enhance the effectiveness of LLM prompts.

\section*{Ethics Statement}
This work uses only publicly available benchmark datasets, including HotpotQA, 2WikiMultihopQA, and MuSiQue, and does not involve the collection of personally identifiable or sensitive information, human-subject experiments, or user studies. All datasets and pretrained models are used solely for academic research and evaluation purposes in accordance with their intended licenses and usage conditions.

\bibliography{custom}

\appendix

\section{Dataset Details}
To comprehensively evaluate the multi-hop reasoning and knowledge structuring capabilities of ThreadRAG, we conduct our experiments across three standard benchmarks. Detailed descriptions of each dataset are provided below:

\textbf{HotpotQA} \cite{yang2018hotpotqa} is a widely-used, crowd-sourced dataset derived from English Wikipedia, comprising approximately 113k questions. It specifically requires models to bridge information across the introductory paragraphs of multiple distinct articles to deduce the correct answer. While the original dataset provides golden paragraphs and human-annotated supporting sentences, we evaluate our framework strictly under a realistic open-domain setting. In this setting, golden contexts are not provided \textit{a priori}; instead, the system must retrieve and structure relevant information from a massive corpus, which rigorously tests the robustness of our Semantic Topology Graph (STG) against retrieval noise and distractors.

\textbf{2WikiMultihopQA (2Wiki)} \cite{ho2020constructing} focuses on evaluating complex 2-hop reasoning, placing a strong emphasis on compositional deduction and comparative analysis. The questions in this dataset are systematically constructed by combining both structured knowledge (from Wikidata) and unstructured texts (from Wikipedia). This structural heterogeneity makes it an ideal testbed for evaluating whether our framework can seamlessly align and linearize varied knowledge formats into a cohesive, modality-native narrative flow without semantic loss.

\textbf{MuSiQue} \cite{trivedi2022musique} is a highly demanding dataset carefully constructed by composing bottom-up single-hop queries into intricate multi-hop reasoning chains. Compared to HotpotQA and 2Wiki, MuSiQue poses a significantly greater challenge as it features deep 2- to 4-hop questions and is meticulously designed to prevent reasoning shortcuts. This dataset effectively stress-tests the capability of our Energy-Driven Pathfinding (EDP) mechanism to maintain long-range logical coherence and prevent structural truncation in extremely deep reasoning scenarios.

\section{Baseline Descriptions}
\label{sec:baselines}

To comprehensively evaluate the performance of ThreadRAG and clearly demonstrate the characteristics of different knowledge representation paradigms, we benchmark our framework against five distinct categories of state-of-the-art baselines.

\textbf{Non-Graph RAG.} This category encompasses text-centric paradigms that tackle multi-hop reasoning over plain text without relying on explicit graph structures. Instead of using structural priors, these approaches lean heavily on the LLM's intrinsic capacity to process continuous contexts and deduce logical chains. To evaluate this paradigm, we select several prominent baselines: \textbf{RA-CoT} \cite{wei2022chain} and \textbf{ReAct} \cite{yao2022react}, which enhance standard text reasoning by integrating retrieval with chain-of-thought and acting, respectively; \textbf{IRCoT} \cite{trivedi2023interleaving} and \textbf{Iter-RetGen} \cite{shao2023enhancing}, which build continuous reasoning chains by iteratively interleaving retrieval steps with generation; and \textbf{Self-Ask} \cite{press2023measuring}, which structures the context by decomposing complex queries into manageable textual sub-queries.

\textbf{KG-Chunk RAG.} Methods in this paradigm aim to mitigate the semantic fragmentation caused by fine-grained triplet extraction by treating larger text chunks as the fundamental nodes in the knowledge graph. This approach provides structural connectivity while preserving local contextual nuances. We evaluate against \textbf{HippoRAG} \cite{gutierrez2024hipporag}, a neurobiologically inspired framework simulating human long-term memory, and \textbf{LightRAG} \cite{guo2024lightrag}, which focuses on simple, fast, and lightweight graph-based retrieval.

\textbf{KG-Rewriting RAG.} Recognizing the modality mismatch between discrete graph structures and the continuous text sequences LLMs are pre-trained on, this category attempts to bridge the gap through generative rewriting. These methods retrieve relevant subgraphs and employ LLMs to translate them into natural language narratives prior to final reasoning. Representative baselines include \textbf{CoTKR} \cite{wu2024cotkr}, which enhances knowledge rewriting via chain-of-thought, and \textbf{EFSUM} \cite{ko2024evidence}, which focuses on generating evidence-focused summaries from graph structures.

\textbf{KG-Path RAG.} This paradigm provides LLMs with explicit, step-by-step structural guidance by forcing them to reason along discrete relational paths extracted from the knowledge graph. The selected baselines are Think-on-Graph (\textbf{ToG}) \cite{sun2024think} and Paths-over-Graph (\textbf{PoG}) \cite{tan2025paths}. For fair comparison, we adapt ToG and PoG to use the same local KG as LightRAG, replacing their external KGs.

\textbf{KG-Triplet RAG.} Representing the most fine-grained end of the structural extraction spectrum, these methods distill documents into rigid entity-relation triplets and use them as discrete prompts. While offering high-density factual associations, this paradigm often strips away the original narrative context entirely. We compare against \textbf{ERA-CoT} \cite{liu2024era}, which improves reasoning through entity relationship analysis, and \textbf{KnowTrace} \cite{li2025knowtrace}, a robust iterative framework guided by structured knowledge tracing.

\section{Example}

To provide a concrete illustration of the end-to-end reasoning process in ThreadRAG, we present a representative example from MuSiQue in Table~\ref{tab:case_study}. The example traces the complete pipeline, including query analysis and retrieval, semantic node construction, energy-driven evidence selection and topological linearization, and final answer synthesis. It demonstrates how ThreadRAG progressively transforms initially retrieved passages into a compact, context-preserving evidence sequence that supports the final multi-hop reasoning process.

\label{sec:Example}

\begin{table*}[t]
\centering
\small
\renewcommand{\arraystretch}{1.2}
\begin{tabular}{p{0.95\textwidth}}
\toprule
\textbf{Question:} Who sings home alone tonight with the singer of you can crash my party anytime? \\
\midrule
\multicolumn{1}{c}{\textbf{[Phase 1: Query Analysis \& Retrieval]}} \\
\texttt{\# Entity Extraction:} \\
- Entities to retrieve: "Home Alone Tonight", "Crash My Party" \\
\texttt{\# Retrieve Relevant Textual Passages:} \\
- \textit{Take Me Home Tonight (song):} "Take Me Home Tonight" is a song by American rock singer Eddie Money... \\
- \textit{Luke Bryan discography:} American country music singer and songwriter Luke Bryan has released... \\
- \textit{Crash My Party (song):} "Crash My Party" is a song recorded by American country artist Luke Bryan... \\
- \textit{Home Alone Tonight:} "Home Alone Tonight" is a song recorded by American country music artist... \\

\\
\multicolumn{1}{c}{\textbf{[Phase 2: Semantic Topology Mapper (STM)]}} \\
\texttt{\# Semantic Node Extraction:} \\
- \textbf{SN 1:} "Take Me Home Tonight" is a song by American rock singer Eddie Money. \\
- \textbf{SN 2:} It was released in August 1986 as the lead single from his album "Can't Hold Back". \\
\texttt{\# Semantic Topology Graph (STG) Construction:} \\
- Graph initialized with 270 context-intact nodes and 878 structural (narrative/semantic/lexical) edges. \\

\\
\multicolumn{1}{c}{\textbf{[Phase 3: Energy-Driven Pathfinding (MAP)]}} \\
\texttt{\# Anchor Identification \& Energy Diffusion:} \\
- 5 query-relevant anchor nodes identified based on the extracted entities. \\
\texttt{\# Graph Pruning \& Topological Linearization:} \\
- 9 highly activated nodes retained out of 270 after energy threshold pruning. \\
- \textit{Linearized Context (Modality-Native Prompt):} \\
  (1) "Crash My Party" is a song written by Rodney Clawson and Ashley Gorley, and recorded by American country artist Luke Bryan. \\
  (2) Crash My Party is the fourth studio album by American country music artist Luke Bryan. \\
  (3) "Home Alone Tonight" is a song recorded by American country music artist Luke Bryan as a duet with Karen Fairchild of American country music group Little Big Town for his fifth studio album, "Kill the Lights" (2015). \\
  (4) ...\\

\\
\multicolumn{1}{c}{\textbf{[Phase 4: Agent Evaluation \& Answer Synthesis]}} \\
\texttt{\# Reasoning Thought Process:} \\
- Step 1: Identify "the singer of you can crash my party anytime". From the linearized context, "Crash My Party" is a song recorded by Luke Bryan. \\
- Step 2: Identify who sings "Home Alone Tonight" with Luke Bryan. From the context, "Home Alone Tonight" is a duet with Karen Fairchild. \\
- Step 3: Therefore, the person who sings "Home Alone Tonight" with Luke Bryan is Karen Fairchild. \\
\textbf{Answer:} Karen Fairchild \\
\bottomrule
\end{tabular}
\caption{Case on MuSiQue.}
\label{tab:case_study}
\end{table*}

\section{Implementation Details of the Semantic Segmentation Model}
\label{sec:segmentation_details}

In response to the necessity for high-fidelity, context-intact partitioning during the Semantic Topology Mapper (STM) phase, we explicitly detail the architecture, data curation, and training pipeline of the MLP-based boundary predictor (introduced in Equation 2).

\textbf{Encoder Architecture and Inference Pipeline.} The ``lightweight encoder'' utilized in our framework is implemented using the pre-trained \texttt{RoBERTa-base} model. To prepare the inputs, raw continuous documents are first segmented into natural sentences using the NLTK library. To maximize contextual awareness while strictly adhering to the 512-token limit of the RoBERTa architecture, we dynamically pack consecutive sentences into continuous blocks of under 512 tokens. The model processes these text blocks to extract dense representations, simultaneously evaluating the semantic continuity at the boundaries between all adjacent sentences within the chunk. This batch-processing approach provides a robust foundation for capturing localized semantic shifts without incurring the massive latency associated with using LLMs for real-time inference. During generation, a semantic breakpoint is formally triggered if the predicted boundary probability $p_i$ exceeds a confidence threshold of $\theta = 0.65$.

\textbf{Data and Supervision Signal (Silver Labels).} To circumvent the prohibitive costs of human annotation while maintaining high segmentation quality, we employ a Knowledge Distillation (KD) paradigm using pseudo-labels (Silver Labels). Specifically, we sample 100 complex multi-hop documents from the Wikipedia corpus. We then prompt powerful teacher models (e.g., GPT-4o-mini and Qwen3.5-Plus) to perform rigorous "Semantic Node Partitioning" on this subset, instructing them to split the texts along natural narrative breaks. This zero-shot distillation yields a high-quality dataset comprising 1,553 distinct semantic nodes. The binary boundaries generated by these teacher models serve as the positive supervision signals ($y_i=1$ for a narrative boundary, $y_i=0$ for continuous context) to train our lightweight boundary predictor.

\textbf{Training Setup and Loss Function.} The model is optimized using the AdamW optimizer. To ensure stable convergence while fine-tuning, we apply a learning rate of $1 \times 10^{-5}$ for the MLP prediction head. The MLP is optimized using the standard Binary Cross-Entropy (BCE) loss:
\begin{equation}
    \mathcal{L}_{seg} = - \frac{1}{N} \sum_{i=1}^{N} \Big[ y_i \log(p_i) + (1 - y_i) \log(1 - p_i) \Big]
\end{equation}
where $N$ is the total number of evaluated sentence boundaries within the input chunks, $y_i \in \{0, 1\}$ is the distillation pseudo-label, and $p_i$ is the predicted boundary probability derived from the MLP head.

\textbf{Generalizability (Shared Across Datasets).} It is crucial to note that the segmentation model is \textit{not} dynamically re-trained or overfitted to individual evaluation sets. The \texttt{RoBERTa-base} encoder and the MLP head are trained exactly once using the distilled pseudo-labels. The resulting weights are then frozen and uniformly shared across all downstream evaluations (HotpotQA, 2WikiMultihopQA, and MuSiQue). This zero-shot inference approach ensures strict evaluation fairness and proves that our STM possesses strong generalization capabilities across unseen open-domain texts.

\textbf{Prompt for Dataset Generation.} To ensure reproducibility, the exact instruction template provided to the teacher models for generating the silver labels is presented in Table \ref{tab:prompt_dataset_gen}.

\begin{table}[htbp]
\centering
\small
\begin{tabular}{@{}p{0.96\linewidth}@{}}
\toprule
\textbf{Prompt 3: Dataset Generation} \\
\midrule
You are an expert linguistic annotator. Your task is to perform ``Semantic Node Partitioning'' on a given multi-hop document. 
Your goal is to divide the continuous text into self-contained ``Semantic Nodes'' along natural narrative, logical, or topical breaks. \\

\vspace{0.2em}
\textbf{\#\# Rules for Partitioning:} \\
1. \textbf{Contextual Integrity:} Each semantic node should be a coherent unit of information (e.g., a complete description of an entity, a specific event, or a logical step). Do not split sentences that are tightly coupled (e.g., pronouns referring to the immediately preceding sentence). \\
2. \textbf{Granularity:} A node typically consists of 1 to 2 sentences. Avoid creating overly long nodes that merge distinct topics, and avoid overly granular nodes (e.g., splitting a single compound sentence). \\
3. \textbf{Exact Match:} The text of the partitioned nodes MUST perfectly reconstruct the original document when concatenated. Do not alter, add, or remove any words. \\

\vspace{0.2em}
\textbf{\#\# Output Format:} \\
You MUST output a valid JSON array of strings, where each string is a continuous segment (Semantic Node) of the original text. \\

\vspace{0.2em}
\textbf{\#\# Example:} \\
\textit{Original Text:} ``Mickey's Safari in Letterland is a 1993 educational NES video game. It stars Disney's Mickey Mouse. The successor title, Walt Disney's Comics and Stories, began publication with the Oct 1940 issue. It featured the Gottfredson serials as a prominent feature.'' \\

\vspace{0.2em}
\textit{Output:} \\
\texttt{[} \\
\texttt{\ \ "Mickey's Safari in Letterland is a 1993 educational NES video} \\
\texttt{\ \ game. It stars Disney's Mickey Mouse.",} \\
\texttt{\ \ "The successor title, Walt Disney's Comics and Stories, began} \\
\texttt{\ \ publication with the Oct 1940 issue. It featured the} \\
\texttt{\ \ Gottfredson serials as a prominent feature."} \\
\texttt{]} \\

\vspace{0.2em}
\textbf{\#\# Input Document:} \\
\{document\_text\} \\

\vspace{0.2em}
Output ONLY the JSON array: \\
\bottomrule
\end{tabular}
\caption{Prompt Templates of Dataset Generation.}
\label{tab:prompt_dataset_gen}
\end{table}

\section{Additional Ablation and Intrinsic Evaluation}
\label{sec:additional_ablation}

To further address the concerns regarding graph topology, evidence selection, and semantic segmentation, we conduct additional controlled ablations and intrinsic evaluations. These experiments complement the main-text analyses by more explicitly isolating the contributions of graph-based evidence organization, Energy-Driven Pathfinding (EDP), and the semantic boundary predictor.

\subsection{Graph-Free Ablation}
\label{sec:graph_free_ablation}

To more cleanly isolate the contribution of graph-based evidence organization from the remaining components of ThreadRAG, we construct a \textbf{Graph-Free} variant. This variant retains the same retrieval pipeline, iterative reasoning procedure, downstream LLM, prompts, and evidence budget as ThreadRAG, while removing Semantic Topology Graph (STG) construction and EDP. Instead, retrieved passages are selected according to query--passage similarity and concatenated following their original document order.

\begin{table}[ht]
\centering
\small
\setlength{\tabcolsep}{7pt}
\begin{tabular}{lccc}
\toprule
\textbf{Method} & \textbf{HotpotQA} & \textbf{2Wiki} & \textbf{MuSiQue} \\
\midrule
Graph-Free & 0.274 & 0.258 & 0.206 \\
ThreadRAG  & \textbf{0.422} & \textbf{0.372} & \textbf{0.324} \\
\bottomrule
\end{tabular}
\caption{Controlled Graph-Free ablation in terms of Exact Match (EM). The Graph-Free variant preserves retrieval and iterative reasoning while removing graph construction and EDP.}
\label{tab:graph_free_ablation}
\end{table}

As shown in Table~\ref{tab:graph_free_ablation}, removing graph-based evidence organization leads to consistent absolute EM decreases of 0.148, 0.114, and 0.118 on HotpotQA, 2Wiki, and MuSiQue, respectively. These results indicate that the performance gain of ThreadRAG cannot be attributed solely to iterative retrieval or reasoning, and that explicitly organizing retrieved evidence through the semantic topology contributes substantially to multi-hop reasoning performance.

\subsection{Comparison of Subgraph Selection Strategies}
\label{sec:selection_strategy}

We further investigate whether the advantage of EDP arises specifically from its topology-aware evidence propagation rather than merely from selecting a connected subgraph. We compare EDP against two standard graph traversal strategies, breadth-first search (BFS) and depth-first search (DFS), as well as a condition without subgraph-based selection. All variants operate on the same candidate graph and use an identical evidence budget; only the evidence selection strategy is changed.

\begin{table}[ht]
\centering
\small
\setlength{\tabcolsep}{8pt}
\begin{tabular}{lcc}
\toprule
\textbf{Subgraph Strategy} & \textbf{EM} & $\boldsymbol{\Delta}$ \textbf{vs. EDP} \\
\midrule
No subgraph      & 0.286 & $-0.038$ \\
BFS (depth = 3)  & 0.294 & $-0.030$ \\
DFS (depth = 3)  & 0.302 & $-0.022$ \\
\textbf{EDP}     & \textbf{0.324} & -- \\
\bottomrule
\end{tabular}
\caption{Comparison of evidence selection strategies on MuSiQue using LLaMA3-8B-Instruct. All methods use the same candidate graph and evidence budget.}
\label{tab:subgraph_strategy}
\end{table}

Table~\ref{tab:subgraph_strategy} shows that both BFS and DFS underperform EDP, with absolute EM decreases of 0.030 and 0.022, respectively. The no-subgraph condition exhibits an even larger decrease of 0.038. Fixed-depth traversal strategies are constrained by local graph exploration: BFS tends to expand broadly within a limited depth, whereas DFS follows individual structural branches. In contrast, EDP propagates energy jointly from multiple query-relevant anchors over the graph, allowing evidence distributed across distant regions to contribute to the final selection. This property is particularly useful for MuSiQue, where answering a question often requires aggregating evidence across 2--4 reasoning hops.

\subsection{Intrinsic Evaluation of Semantic Segmentation}
\label{sec:segmentation_evaluation}

In addition to the downstream ablation reported in the main paper, we directly evaluate the boundary prediction quality of the semantic segmentation model. We construct an independent held-out evaluation set containing 50 Wikipedia documents that are disjoint from the 100 documents used to train the boundary predictor. Teacher-generated semantic boundaries are used as reference annotations, resulting in 719 reference boundaries in total.

\begin{table}[t]
\centering
\small
\setlength{\tabcolsep}{12pt}
\begin{tabular}{ccc}
\toprule
\textbf{Precision} & \textbf{Recall} & \textbf{F1} \\
\midrule
0.948 & 0.861 & 0.902 \\
\bottomrule
\end{tabular}
\caption{Intrinsic evaluation of the semantic boundary predictor on 50 held-out Wikipedia documents. Scores measure agreement with teacher-generated reference boundaries.}
\label{tab:segmentation_intrinsic}
\end{table}

As shown in Table~\ref{tab:segmentation_intrinsic}, the lightweight boundary predictor achieves a precision of 0.948, recall of 0.861, and F1 score of 0.902. The high precision indicates that the model rarely introduces unsupported semantic breaks, while the recall suggests that most reference boundaries are successfully identified. Together with the downstream performance reported in the main paper, these results provide complementary evidence that the learned segmentation model can preserve coherent semantic units without relying on dataset-specific retraining.

\section{Prompt Templates}
\label{sec:prompt_templates}

In this section, we detail the core prompt templates utilized in ThreadRAG. To ensure reproducibility, we present the exact instructions provided to the underlying Large Language Models for both entity extraction and the iterative information sufficiency evaluation.

\subsection{Entity Extraction Prompt}
This prompt (Table \ref{prompt:Entity Extraction}) is designed to instruct the LLM to extract key reasoning entities from the user's multi-hop query, stripping away generic stop words to form precise search anchors.

\begin{table}[t]
\centering
\small
\renewcommand{\arraystretch}{1.2}
\begin{tabular}{@{}p{0.95\linewidth}@{}}
\toprule
\textbf{Prompt 1: Entity Extraction} \\
\midrule
\ttfamily \raggedright \arraybackslash
You are an entity extraction assistant. Extract key entities from the question for knowledge retrieval. \\
\vspace{1ex}
Rules: \\
1. Extract only key entities (persons, places, organizations, products, etc.) \\
2. Return ONLY a JSON array of entity strings \\
3. DO NOT include generic words like "who", "what", "where", "the", "a", etc. \\
4. For multi-hop questions, include ALL entities needed for reasoning \\
5. Keep entity names concise (2-4 words max) \\
\vspace{1ex}
Examples: \\
Question: "Who directed Inception?" \\
Entities: ["Inception"] \\
\vspace{0.5ex}
Question: "Aside from the Apple Remote, what other device can control Front Row?" \\
Entities: ["Apple Remote", "Front Row"] \\
\vspace{0.5ex}
Question: "What sport did the person who founded Google play in college?" \\
Entities: ["Google", "Larry Page"] \\
\bottomrule
\caption{Prompt Templates of Entity Extraction.}
\label{prompt:Entity Extraction}
\end{tabular}
\end{table}

\subsection{Information Sufficiency Discriminator Prompts}
To prevent premature hallucination, ThreadRAG employs an Information Sufficiency Discriminator. As detailed in Table \ref{prompt:Discriminator}, this system prompt establishes strict evaluation criteria, guiding the LLM to carefully assess whether the provided context contains adequate factual information to fully resolve the user's query before proceeding to the final answer generation.

\begin{table}[t!]
\centering
\small
\renewcommand{\arraystretch}{1.2}
\begin{tabular}{@{}p{0.95\linewidth}@{}}
\toprule
\textbf{Prompt 2: Discriminator} \\
\midrule
\ttfamily \raggedright \arraybackslash
You are an Information Sufficiency Discriminator. \\
Your task is to evaluate whether the provided context is sufficient to answer the user's question. \\
\vspace{1ex}
\#\# Output Format (JSON): \\
\{"status": "Sufficient" or "Insufficient", "reason": "brief explanation if insufficient, empty string if sufficient"\} \\
\vspace{1ex}
\#\# Evaluation Criteria: \\
- If the context contains enough factual information to answer the question, output \{"status": "Sufficient", "reason": ""\} \\
- If the context is missing key information, output \{"status": "Insufficient", "reason": "what specific information is missing"\} \\
- Be strict: only output "Sufficient" if you are confident the question can be fully answered from the context. \\
\bottomrule
\caption{Prompt Templates of Discriminator.}
\label{prompt:Discriminator}
\end{tabular}
\end{table}

\section{Qualitative Error Analysis}
\label{sec:qualitative_error_analysis}

\begin{figure}[t]
\centering
\includegraphics[width=0.5\textwidth]{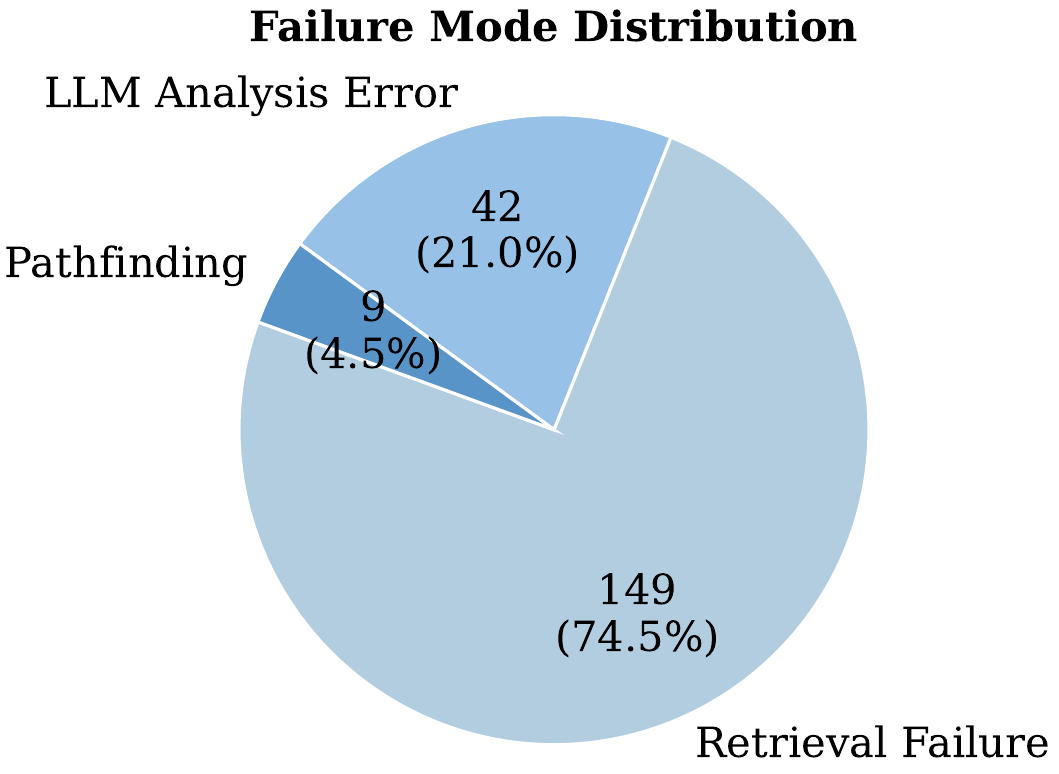}
\caption{Distribution of failure modes analyzed from 200 incorrect predictions.}
\label{fig:graph_effect1}
\end{figure}

To systematically investigate the boundary conditions and inherent limitations of the ThreadRAG framework, we conducted a rigorous qualitative error analysis on the HotpotQA dataset (in Figure \ref{fig:graph_effect1}). Specifically, we randomly sampled 200 incorrect prediction cases from the evaluation pool and manually scrutinized their underlying reasoning traces. These failures were comprehensively classified into three distinct categories based on their root causes:
\begin{itemize}[leftmargin=*, itemsep=0pt, parsep=0pt, topsep=0pt]
    \item \textbf{Retrieval Failure (74.5\%, 149 cases):} This dominant failure mode occurs when the upstream initial retriever fails to surface any documents containing the gold supporting evidence within the top-$k$ retrieval pool. Consequently, the subsequently constructed Semantic Topology Graph (STG) lacks the necessary factual foundation, rendering correct deduction impossible.
    \item \textbf{LLM Analysis Error (21.0\%, 42 cases):} In this scenario, the ThreadRAG pipeline functions flawlessly, successfully linearizing a perfect, context-intact prompt that encapsulates the complete multi-hop reasoning chain. However, the downstream baseline Large Language Model suffers from intrinsic logical deficiencies or hallucinations during the final step, failing to deduce the correct target answer despite having access to the pristine context.
    \item \textbf{Pathfinding Error (4.5\%, 9 cases):} This represents the internal boundary condition of our system. During exceptionally long-range or complex multi-hop path exploration, the energy propagated via our Energy-Driven Pathfinding (EDP) module decays beyond acceptable margins, or falls slightly below the rigid structural pruning threshold. As a result, critical evidence nodes are inadvertently truncated or filtered out during the topological linearization phase.
\end{itemize}

This empirical breakdown of failure modes strongly validates the efficacy, robustness, and architectural soundness of our proposed Energy-Driven Pathfinding (EDP) mechanism. The statistical distribution clearly illustrates that a staggering 95.5\% of the total failure cases are exclusively triggered by orthogonal, external bottlenecks—namely, the hard recall limits of the upstream retriever and the intrinsic reasoning ceilings of the baseline LLM. In sharp contrast, errors stemming from our internal graph pathfinding and structural pruning components constitute an absolute minority (merely 4.5\%). These quantitative insights demonstrate that the EDP module successfully navigates complex knowledge topologies to eliminate redundant context noise while maintaining an exceptionally high fidelity in preserving critical factual chains for downstream generation.

\end{document}